\documentclass[letterpaper, 10 pt, conference]{ieeeconf}  % Comment this line out

\IEEEoverridecommandlockouts                              % This command is only
\usepackage[colorlinks,linkcolor=green,citecolor=red,urlcolor=blue,bookmarks=true,hypertexnames=true]{hyperref} 
\usepackage[utf8]{inputenc}
\usepackage[T1]{fontenc}

\usepackage{graphicx} % for pdf, bitmapped graphics files
\usepackage{epsfig} % for postscript graphics files
\usepackage{mathptmx} % assumes new font selection scheme installed
\usepackage{mathptmx} % assumes new font selection scheme installed
\usepackage{amsmath} % assumes amsmath package installed
\usepackage{amssymb}  % assumes amsmath package installed
\usepackage{booktabs}
\usepackage{url}
\usepackage{multirow}
\usepackage{multicol}
\usepackage{diagbox}
\usepackage{graphicx}
\usepackage{makecell} % 实现表格居中宏包
\usepackage{balance}
\usepackage{tabularx}
\usepackage{threeparttable}
\title{\LARGE \bf
Multimodal HD Mapping for Intersections by Intelligent Roadside Units}

\author{Zhongzhang Chen$^{2}$, Miao Fan$^{1,*}$, 
Shengtong Xu$^{3}$, Mengmeng Yang$^{4}$, Kun Jiang$^{4}$, Xiangzeng Liu$^{5}$, Haoyi Xiong$^{6}$ % <-this % stops a space
\\
\\
$^{1}$Beijing Institute of Graphic Communication,
$^{2}$NavInfo Co. Ltd.,
$^{3}$Autohome Inc.,
\\
$^{4}$Tsinghua University, $^{5}$Xidian University, $^{6}$Baidu Inc.
\thanks{*Corresponding author: Miao Fan (miao.fan@ieee.org), professor at Beijing Institute of Graphic Communication, senior member of IEEE.}
}
        
\begin{document}

\maketitle
\thispagestyle{empty}
\pagestyle{empty}

%%%%%%%%%%%%%%%%%%%%%%%%%%%%%%%%%%%%%%%%%%%%%%%%%%%%%%%%%%%%%%%%%%%%%%%%%%%%%%%%
\begin{abstract}

High-definition (HD) semantic mapping of complex intersections poses significant challenges for traditional vehicle-based approaches due to occlusions and limited perspectives. This paper introduces a novel camera-LiDAR fusion framework that leverages elevated intelligent roadside units (IRUs). Additionally, we present RS-seq, a comprehensive dataset developed through the systematic enhancement and annotation of the V2X-Seq dataset. RS-seq includes precisely labelled camera imagery and LiDAR point clouds collected from roadside installations, along with vectorized maps for seven intersections annotated with detailed features such as lane dividers, pedestrian crossings, and stop lines. This dataset facilitates the systematic investigation of cross-modal complementarity for HD map generation using IRU data. The proposed fusion framework employs a two-stage process that integrates modality-specific feature extraction and cross-modal semantic integration, capitalizing on camera high-resolution texture and precise geometric data from LiDAR. Quantitative evaluations using the RS-seq dataset demonstrate that our multimodal approach consistently surpasses unimodal methods. Specifically, compared to unimodal baselines evaluated on the RS-seq dataset, the multimodal approach improves the mean Intersection-over-Union (mIoU) for semantic segmentation by 4\% over the image-only results and 18\% over the point cloud-only results. This study establishes a baseline methodology for IRU-based HD semantic mapping and provides a valuable dataset for future research in infrastructure-assisted autonomous driving systems.

\end{abstract}

{\keywords Semantic HD Maps, multimodal data, intersections, intelligent roadside units.}

%%%%%%%%%%%%%%%%%%%%%%%%%%%%%%%%%%%%%%%%%%%%%%%%%%%%%%%%%%%%%%%%%%%%%%%%%%%%%%%%

\section{Introduction}
% old: HD semantic maps play a crucial role in autonomous driving. Semantic features, including lane dividers, stop lines, and pedestrian crossings on HD maps, offer precise information about road location that is essential for autonomous vehicles. With advancements in deep learning technologies, the generation of HD maps through the integration of camera images and LiDAR point clouds collected by vehicles has become increasingly prevalent. Recent approaches~\cite{li2022hdmapnet, liao2023maptrv2,  liu2024mgmap} have explored the fusion of camera images and LiDAR point clouds from vehicles to facilitate the automated production of HD maps. Despite advancements in technical methods, HD map production often still requires substantial manual intervention, particularly at intersections with complex environments and numerous elements, thus constraining production efficiency. The limited viewpoint of the vehicle results in multiple trips to these intersections during data collection, significantly increasing the cost of HD map creation. Furthermore, traditional methods do not promptly update road elements at intersections, restricting the applicability of HD maps in autonomous driving contexts.

Semantic HD maps are essential for autonomous driving, as they provide precise road location details through semantic features such as lane dividers, stop lines, and pedestrian crossings. Recent advancements in deep learning technologies have enabled the automated generation of semantic HD maps by integrating camera images and LiDAR point clouds collected by vehicles. Several studies~\cite{li2022hdmapnet, liao2023maptrv2, liu2024mgmap} have investigated methods for fusing these modalities to facilitate HD map production. However, despite these technological advancements, creating HD maps requires significant manual intervention, especially at intersections with complex environments and numerous road elements. These challenges hinder production efficiency. Additionally, the limited perspective of vehicle-mounted sensors requires multiple data collection trips to the same intersection, further increasing the cost of HD map generation. Traditional approaches also struggle to update road elements at intersections promptly road elements at intersections, reducing the applicability of HD maps in autonomous driving systems that require real-time accuracy.

\begin{figure*}
  \includegraphics[width=\textwidth]{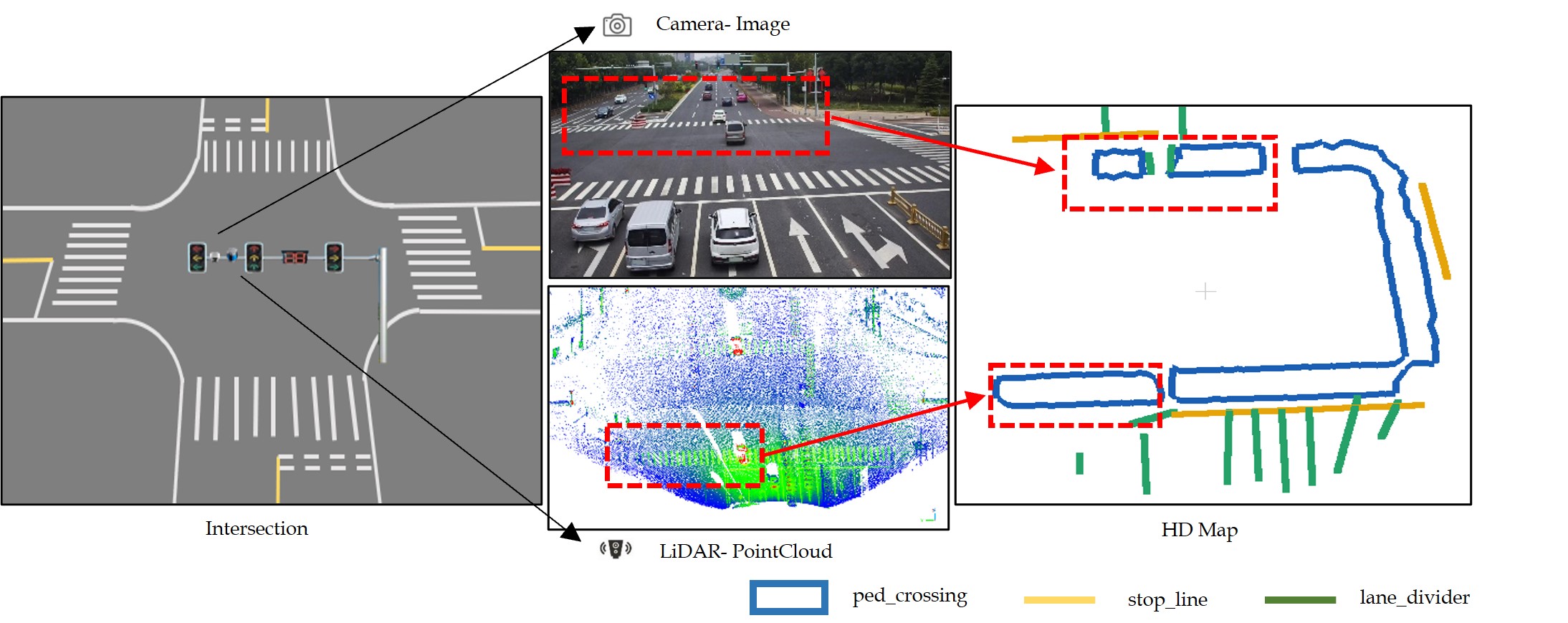}
  \caption{
  %Multimodal data (including camera images and LiDAR point clouds) collections and HD mapping by Intelligent roadside units.
  %
  Multimodal data acquisition and semantic HD map generation framework utilizing intelligent roadside units: integrating camera imagery and LiDAR point clouds for comprehensive semantic mapping.
  }
  \label{fig:1}
\end{figure*}

With the advancement and application of collaborative sensing technologies, efforts have been directed towards incorporating IRUs into creating semantic HD maps at intersections. Positioned at elevated locations, IRUs provide an expanded field of view and reduce obstructions, enabling the generation of more comprehensive semantic HD maps for these critical areas. Relying exclusively on LiDAR point clouds during HD map creation poses challenges such as difficulty in labelling and the absence of texture information. Conversely, exclusive dependence on camera imagery reduces accuracy and increases sensitivity to lighting conditions. Multimodal approaches that integrate LiDAR point clouds with camera imagery leverage the strengths of both modalities, enhancing the completeness and precision of semantic HD maps.

To advance infrastructure-assisted HD mapping research, we systematically enhanced the V2X-seq dataset~\cite{yu2023v2x} by annotating key road elements, including lane dividers, pedestrian crossings, and stop lines, in both camera images and LiDAR point clouds. This enhanced dataset, RS-seq, facilitates the systematic investigation of cross-modal complementarity in semantic HD map generation.

Our findings show that camera images capture high-resolution texture and colour information crucial for detecting lane dividers. At the same time, LiDAR point clouds offer precise geometric data valuable for identifying pedestrian crossings and stop lines. As illustrated in Fig.~\ref{fig:1}, integrating these complementary features significantly improves mapping accuracy. Quantitative evaluations indicate that our multimodal approach consistently surpasses unimodal methods across diverse intersection scenarios. Using the RS-seq dataset, our approach achieves a 4\% improvement in mIoU for semantic segmentation compared to the best-performing unimodal baseline.

The key contributions of this work are summarized as follows:

\begin{itemize}

% \item Based on the V2X-seq dataset, we expanded the annotation information of camera images and LiDAR point clouds to form a multimodal dataset, RS-seq, for constructing HD semantic maps with roadside unit data.

% \item Based on the RS-seq dataset, we explore the multimodal mapping technique for intelligent roadside unit data and establish a baseline method of constructing HD semantic maps based on camera images and LiDAR point clouds.

% \item For the purpose of study, both code and data are available at \url{https://github.com/CZZGIT/MapFusion}.

%

\item We developed the RS-seq dataset, a richly annotated multimodal dataset that extends the V2X-seq dataset by systematically labelling camera images and LiDAR point clouds. This dataset uses HD semantic map construction data from roadside units.

\item Based on the RS-seq dataset, we investigated multimodal mapping methods for roadside data. We proposed a baseline method for generating HD semantic maps by fusing camera and LiDAR data.

\item We release the dataset and source codes at OneDrive\footnote{\url{https://1drv.ms/f/c/76645c25a8914a0b/Eub_ulaT3KhKkiZgj31IKyQBkAtRiUZ9hkBvjghSpEjipg}} to facilitate reproducibility and further research on infrastructure-assisted mapping.

\end{itemize}

\section{Related Work}

HD semantic mapping of urban intersections presents significant challenges for vehicle-based approaches due to occlusions and limited perspectives. Recent research has explored intelligent roadside infrastructure and multimodal sensor fusion as solutions to these limitations. This section reviews advancements in infrastructure-assisted mapping methods, sensor fusion techniques, and intersection-specific datasets.

\begin{figure*}
\centering
\includegraphics[width=\textwidth]{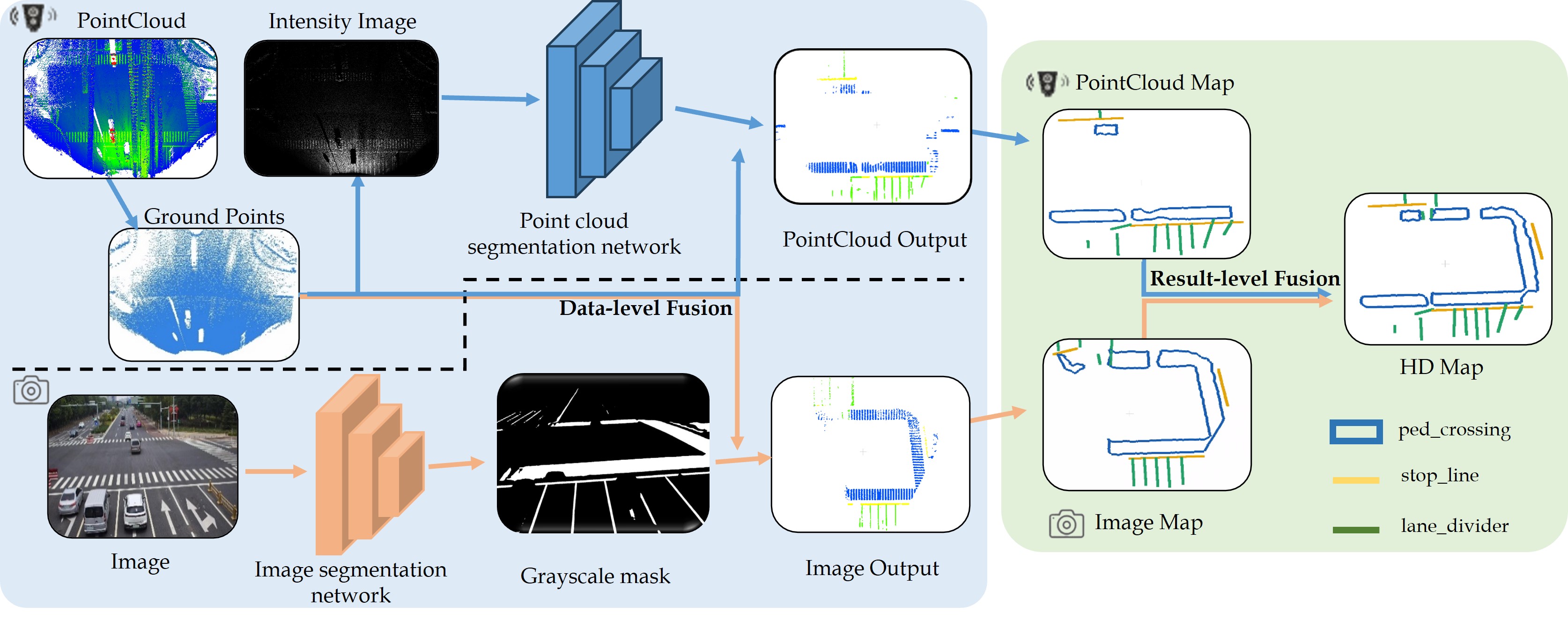}
\caption{%Pipeline overview of the method of creating HD maps based on the roadside units data.
The systematic framework for semantic HD map generation through intelligent roadside unit data processing: a pipeline from multimodal data acquisition to vectorized map representation.
}
\label{fig:2}
\end{figure*}

\subsection{Semantic Map Construction}
In the traditional offline HD map production process, a substantial amount of LiDAR point cloud data is initially collected using vehicle-mounted sensors. This raw data is subsequently processed through Simultaneous Localization and Mapping (SLAM) technology~\cite{lee2020design}, a procedure known for its considerable computational demands and algorithmic complexity, which ensures the accurate alignment and integration of the LiDAR point clouds. Following this, the LiDAR point clouds undergo segmentation and vectorization, and the final map data requires manual validation and correction to ensure accuracy and reliability. The conventional offline mapping approach is characterized by high costs and slow update cycles, rendering it less adaptable and scalable. Recent studies have investigated real-time mapping methods that integrate images from surround-view cameras with LiDAR point clouds to overcome these limitations and lessen dependence on manually annotated semantic maps. Notably, HDMapNet~\cite{li2022hdmapnet} leverages Multi-Layer Perceptrons (MLPs) to map photometric and geometric (PV) features to Bird’s Eye View (BEV) features within the camera branch while employing PointPillars~\cite{Lang_2019_CVPR} to encode BEV features in the LiDAR branch. Similarly, BEVFusion~\cite{liang2022bevfusion} uses Lift-Splat-Shoot (LSS)~\cite{philion2020lift} for view conversion in the camera branch and VoxelNet~\cite{yan2018second} for the LiDAR branch, with integration of these modalities achieved through the BEV alignment module. Superfusion~\cite{10611320} further advances the field by integrating multilevel feature information from camera images and LiDAR point clouds to create long-range HD maps. However, these real-time mapping techniques are constrained by the limited field of view and occlusions from in-vehicle sensors, as well as variations in data quality caused by rapid sensor motion, which can lead to vulnerabilities and incompleteness in current online map construction schemes.

\subsection{Roadside Unit Mapping}
Relying solely on vehicles for road data collection poses limitations due to the vehicle’s restricted field of view. For instance, certain road elements may remain unrecorded at intersections because they are occluded by other vehicles. However, advancements in collaborative perception technology have demonstrated that intelligent roadside units installed at elevated positions can enhance data collected by vehicles. The V2X-ViT framework~\cite{xu2022v2x} utilizes a virtual dataset to validate that LiDAR point cloud data from road test facilities can improve 3D Object detection performance for self-driving vehicles. The researchers proposed a monocular 3D target detection method that utilizes camera images captured by vehicles and intelligent roadside units~\cite{Ye_2022_CVPR}. Research scholars have released the DAIR-V2X~\cite{yu2022dair} and A9~\cite{zimmer2024tumtrafv2x} datasets, derived from real-world sources to facilitate collaborative 3D target detection for vehicles and roads. They have experimented with multimodal fusion modelling for collaborative 3D target detection and tracking tasks, demonstrating the advantages of roadside equipment in enhancing autonomous vehicles’ perception capabilities and reliability. Despite these advancements, research on leveraging intelligent roadside unit data for HD map production remains limited. VI-Map~\cite{he2023vi} represents the first system to integrate data from roadside units and vehicles, achieving real-time HD map construction by extracting BEV features and vectorized map representations. This system significantly improves map accuracy and coverage for autonomous driving. However, VI-Map relies solely on LiDAR point clouds, which lack texture information and experience reduced point density at greater distances. This paper introduces the first HD map creation method using camera images and LiDAR point clouds collected by intelligent roadside units, exploring the comprehensive utilization of such data to produce HD maps and demonstrate promising results.

\subsection{Roadside Unit Dataset}
% V2XSet~\cite{xu2022v2x} employs the high-fidelity CARLA simulator~\cite{dosovitskiy2017carla} along with the OpenCDA cooperative driving automation simulation tool to specifically tackle the challenges posed by real-world noise in V2X communication. V2X-Seq~\cite{yu2023v2x} represents the first extensive continuous V2X dataset, providing a rich array of traffic participant behavior data to support autonomous driving decision-making and safety research. This dataset introduces novel tasks and benchmarks for vehicle-infrastructure cooperation (VIC) in autonomous driving. Additionally, Rope3D~\cite{Ye_2022_CVPR}, also known as the Roadside Perception 3D dataset, aggregates image data from various carriers, scenes, and cameras. The Tumtraffic-v2x dataset~\cite{zimmer2024tumtrafv2x} includes a substantial collection of labeled camera images and LiDAR point clouds from multiple roadside units and vehicle-mounted sensors. We developed the RS-seq dataset based on the publicly available V2X-Seq dataset, incorporating extensive optimizations. This dataset comprises point cloud and image data collected by roadside devices and is meticulously labeled to generate vectorized maps of seven intersections. Key elements such as lane dividers, pedestrian crossings, and stop lines are included in the maps with precision.

V2XSet~\cite{xu2022v2x} leverages the high-fidelity CARLA simulator~\cite{dosovitskiy2017carla} and the OpenCDA tool to address real-world V2X communication noise. V2X-Seq~\cite{yu2023v2x}, the first extensive continuous V2X dataset, offers rich traffic behaviour data for autonomous driving decisions and safety research, introducing new vehicle-infrastructure cooperation(VIC) tasks and benchmarks. Rope3D~\cite{Ye_2022_CVPR} compiles images from diverse sources, while Tumtraffic-v2x~\cite{zimmer2024tumtrafv2x} provides numerous labelled camera images and LiDAR point clouds. 

The RS-seq dataset, an optimized extension of V2X-Seq, comprises synchronized camera images and LiDAR point clouds from multiple intersections collected by IRUs. It includes detailed annotations for key road elements such as lane dividers, pedestrian crossings, and stop lines, enabling the generation of vectorized maps for seven distinct intersections. The generated semantic HD maps in the RS-seq dataset are based on a global coordinate system, but for data security, the coordinates are offset. The dataset was collected from the Beijing Yizhuang area. The frame rate for LiDAR data acquisition is 10 Hz, while the resolution of the camera images is 1920×1080 pixels.
Temporal synchronization between the camera and LiDAR sensor streams is crucial to ensure effective multimodal fusion. The synchronization was achieved through post-processing based on nearest-neighbour timestamp matching, with a maximum tolerance of 50 ms. Any data pairs exceeding this tolerance were excluded to maintain temporal alignment integrity.
The RS-seq dataset was entirely allocated to the test set for final performance reporting.

\begin{table}[h]
\caption{Roadside Unit Multimodal and Unimodal Semantic HD Map Quality Assessment Comparison.}
\begin{center}
\begin{tabular}{c|c|c|c|c}
\toprule
{\textbf{ID}} & {\textbf{Truth}} & {\textbf{Image}} & {\textbf{PointCloud}} & {\textbf{Multimodal}}                                       \\ 
\midrule
\makecell[c]{Intersection \\ \\ No.05 \\ ~~} & \includegraphics[width=0.5in]{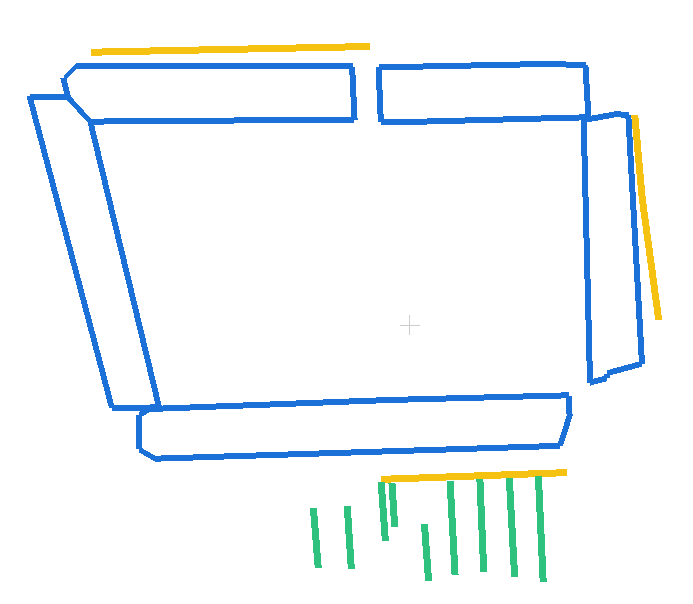}  & \includegraphics[width=0.5in]{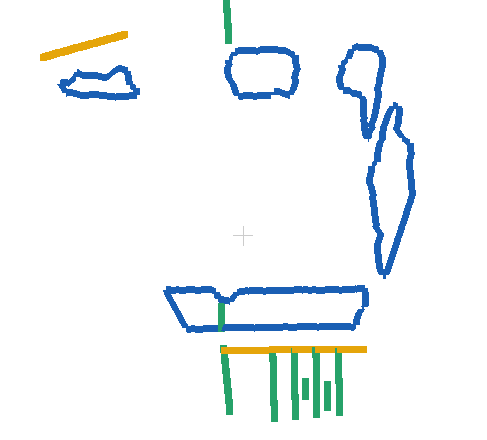} & \includegraphics[width=0.5in]{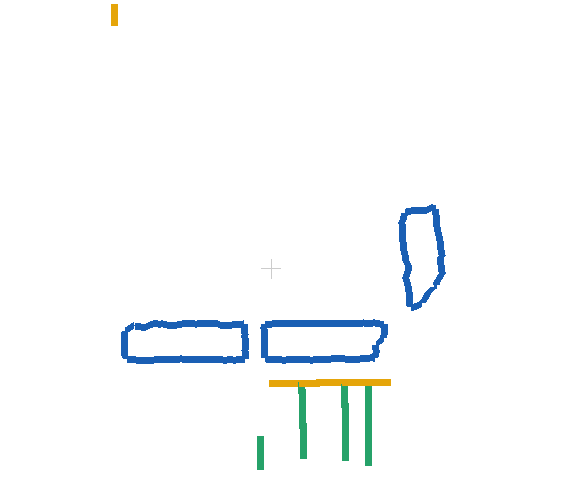} & \includegraphics[width=0.5in]{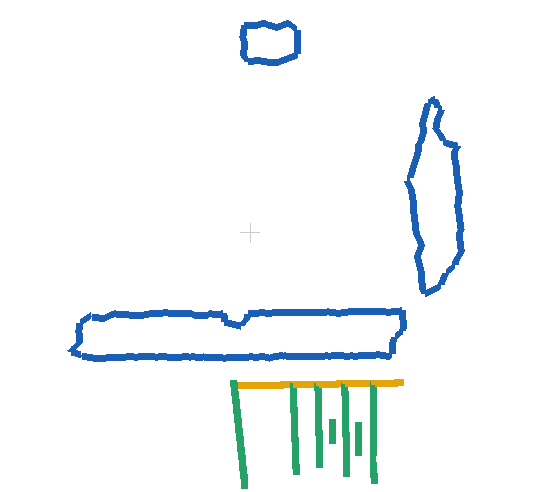} \\
\midrule
\makecell[c]{Intersection \\ \\ No.08 \\ ~~} & \includegraphics[width=0.5in]{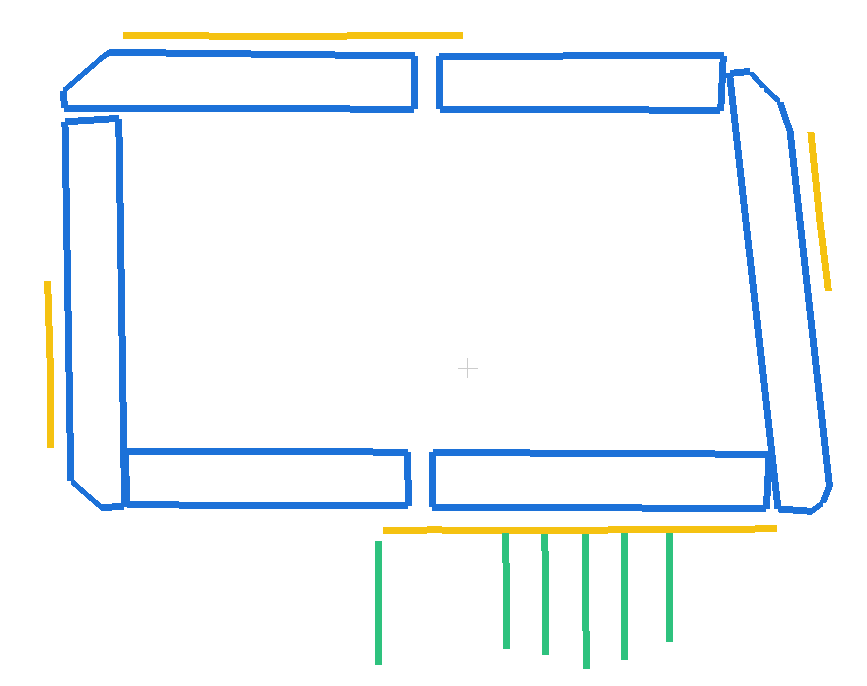}  & \includegraphics[width=0.5in]{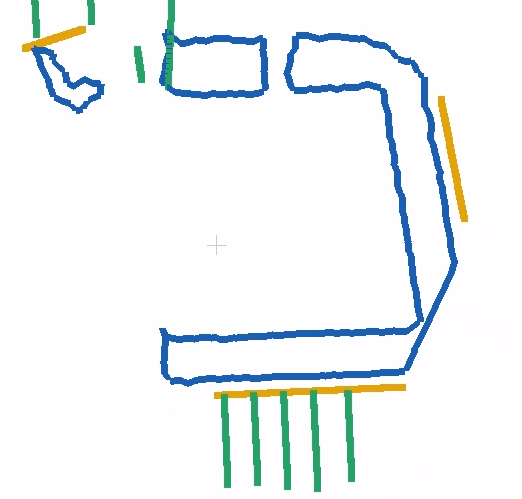} & \includegraphics[width=0.5in]{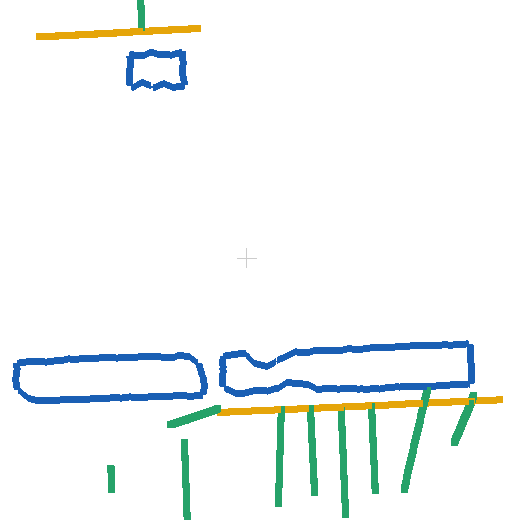} & \includegraphics[width=0.5in]{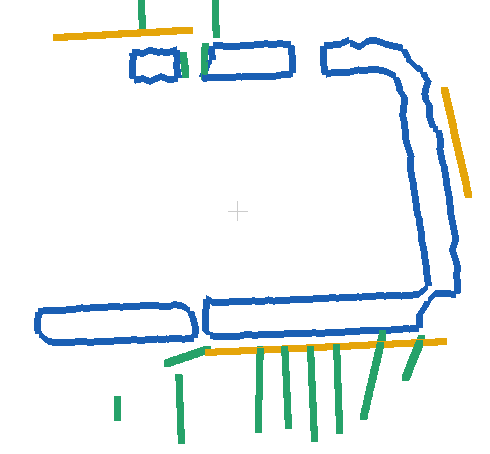} \\
\midrule
\makecell[c]{Intersection \\ \\ No.10 \\ ~~} & \includegraphics[width=0.5in]{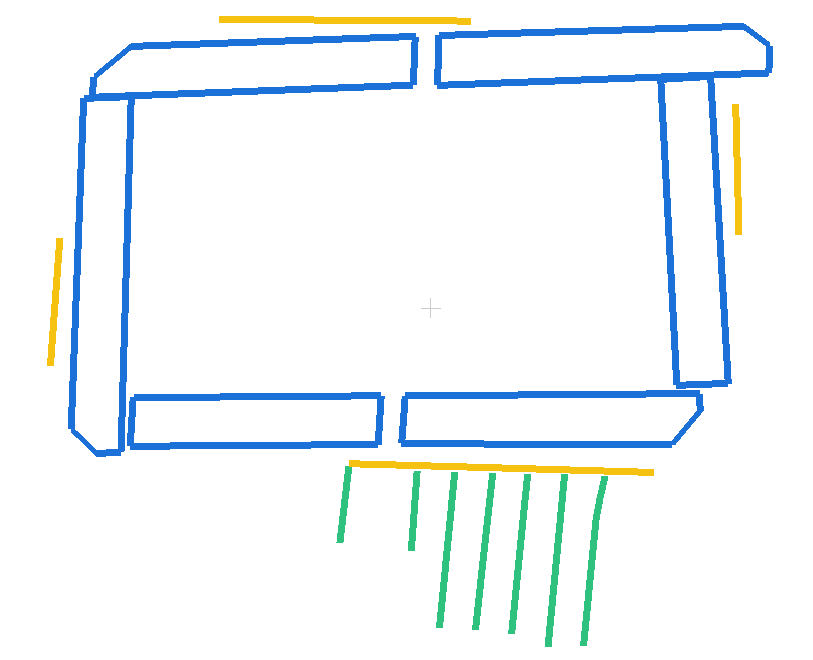}  & \includegraphics[width=0.5in]{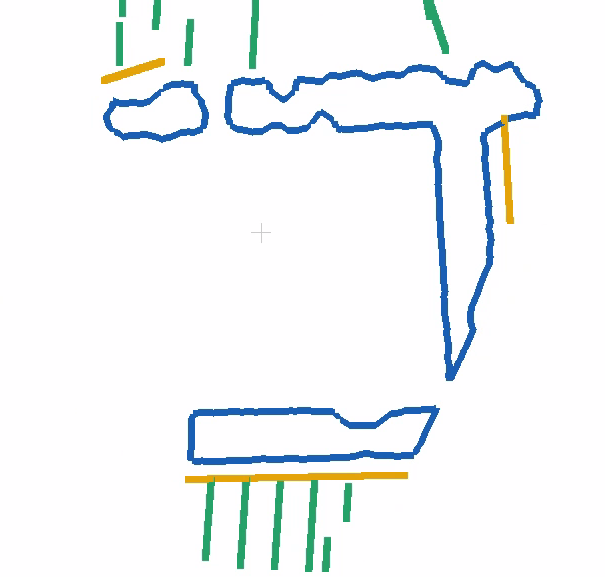} & \includegraphics[width=0.5in]{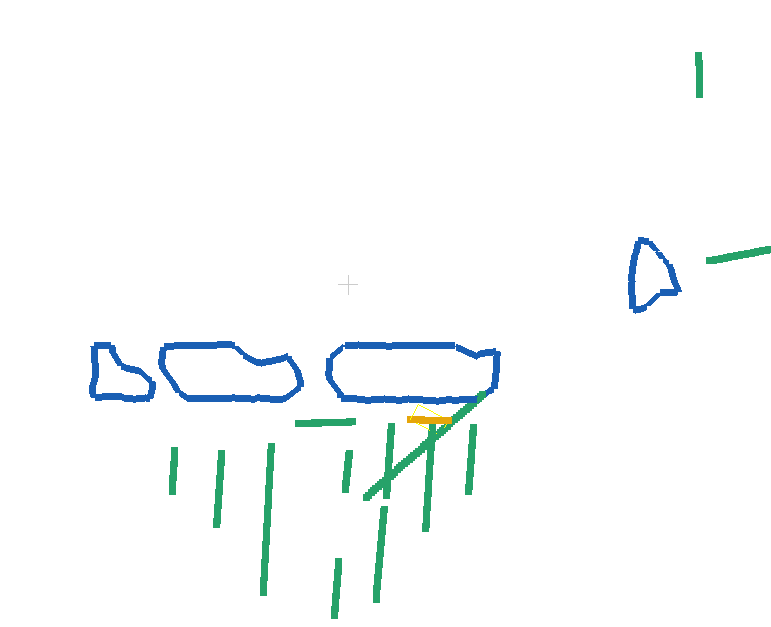} & \includegraphics[width=0.5in]{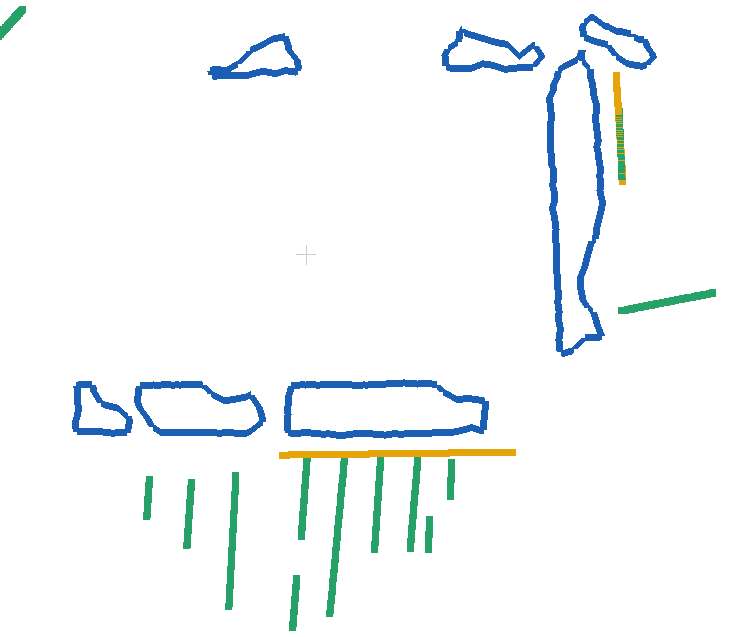} \\
\midrule
mIoU & 1.000 & 0.431 & 0.293 & 0.474 \\
\bottomrule
\end{tabular}
%}
\end{center}
\begin{tablenotes}
\scriptsize
\item {Yellow line segments represent stop lines.
Green line segments represent lane markings.
Blue boxes represent crosswalks.}
\end{tablenotes}

\label{tab:1}
\end{table}

\section{Benchmark Methods}

% Using roadside units to capture images and point cloud data for mapping, we investigate a baseline approach that is a multimodal algorithm using data-level fusion and result-level fusion.

This section presents a baseline multimodal framework for Semantic HD map generation through hierarchical fusion of roadside sensor data. The proposed methodology integrates camera imagery and LiDAR point clouds through both data-level and result-level fusion strategies.

\subsection{Image and Point Cloud Fusion}

Based on the RS-seq dataset, we developed our baseline processing method, a multimodal mapping approach that leverages image and point cloud data. This method incorporates data-level fusion and result-level fusion to fully exploit the advantages of both data types in creating semantic HD maps. The data-level fusion employs an image segmentation network to segment the image, based on the image segmentation results, we extract the class label for each pixel in the image, perform one-hot encoding on the class labels, and generate grayscale mask images according to the encoded labels and their corresponding pixel positions. On the point cloud side, the RANSAC plane fitting method identifies the ground point cloud data. Subsequently, the internal and external parameters of the acquisition equipment are applied to transform the ground point cloud data into pixel coordinates in the image coordinate system, as shown in Formula~\ref{eq:1}. The extracted pixel values corresponding to the ground point cloud data derived from the image segmentation results serve as candidate points. The method of fusing image segmentation results with point cloud ground points is a manifestation of data-level fusion.

\begin{equation}
\begin{aligned}
\label{eq:1}
{Z_C}\left[ \begin{array}{l}
u\\
v\\
1
\end{array} \right] = \left[ {\begin{array}{*{10}{c}}
{\frac{1}{{{d_x}}}}&0&{{u_0}}\\
0&{\frac{1}{{{d_y}}}}&{{v_0}}\\
0&0&1
\end{array}} \right] \cdot \left[ {\begin{array}{*{10}{c}}
f&0&0&0\\
0&f&0&0\\
0&0&1&0
\end{array}} \right] \cdot \\
\left[ {\begin{array}{*{10}{c}}
R&T\\
{{0^T}}&1
\end{array}} \right] \cdot \left[ {\begin{array}{*{10}{c}}
{{X_{\rm{L}}}}\\
{{Y_L}}\\
{{Z_L}}\\
1
\end{array}} \right]
\end{aligned}
\end{equation}
where ${X_L}$, ${Y_L}$, ${Z_L}$ are the world coordinate system coordinates of the point, R and T are the extrinsic parameters of the camera, f is the focal length of the camera, ${u_0}$ and ${v_0}$ are the intersection coordinates of the camera centre and the image, ${d_x}$ and ${d_y}$ are the length and width of a single pixel in the image plane, and $u$ and $v$ are the pixel coordinates of the point.

The point cloud data is segmented, and after the ground point cloud data is gridded, the average reflective intensity of each point in each grid is calculated. Each grid is then treated as a pixel, with the average reflective intensity of the points in the grid used as the pixel’s grey value. This method generates a point cloud intensity image of the ground. The segmentation result of the point cloud intensity image is then obtained using an image segmentation algorithm. Subsequently, the point cloud data corresponding to the pixel values in the grid are extracted to obtain the segmented point cloud data. As shown in Table~\ref{tab:2}, we used several different network models in the image segmentation network to verify the impact of the segmentation network model on the results. While elemental segmentation for the image and point cloud intensity image uses different networks, their network structures are the same.

\begin{equation}\label{eq:2}
\begin{split}
{x_i} = \left\lfloor {\frac{{({x_p} - {X_{\min }})}}{{{X_{size}}}}} \right\rfloor \\
{y_i} = \left\lfloor {\frac{{({y_p} - {X_{\min }})}}{{{Y_{size}}}}} \right\rfloor
\end{split}
\end{equation}
where ${x_i}$ and ${y_i}$ are the indices of the grid where the current point is located, ${X_{min}}$ and ${Y_{min}}$ are the minimum values in the x and y directions of the current point cloud, ${x_p}$ and ${y_p}$ are the coordinate values of the current point, and ${X_{size}}$ and ${y_{size}}$ are the sizes of the grid.

\begin{equation}\label{eq:3}
G(r,c) = \sum\limits_{i = 0}^N {{T_i}} {\rm{/}}N
\end{equation}
where $G(r, c)$ represents the grayscale values of the pixels at the $r$th row and $c$-th column of the grayscale image, $N$ is the number of point clouds in that grid, and ${T_i}$ is the intensity value of the points in that grid.

The point cloud data generated from the preceding two steps are integrated into the result-level fusion process. Subsequently, the point cloud data for each element are clustered and denoised to produce the final point cloud output.

\subsection{Map Vectorization}
The segmentation results of each element obtained from point cloud data segmentation are not directly applied to the map but need to go through a final vectorization step. We follow the specifications of the semantic HD map and vectorize lane markings and stop lines as lines and pedestrian crossings as polygons. Specifically, the point cloud data are denoised using statistical outlier removal (SOR) to remove noise points~\cite{2020Denoising} and then clustered using the nearest neighbour clustering method. For each point cloud cluster of each polygonal element, the $\alpha$-shape~\cite{1994Three} algorithm is used to calculate the edge points of the point cloud clusters. Then, the edge points are connected into a line to get the final vectorization result. For the line element, the point cloud clusters are fitted using the least-squares fitting method~\cite{york1966least}, and the fitted line is obtained as the vectorization result of the component. Finally, the vector map output from the point cloud and the vector map output from the image are fused by taking the concatenation set, thus realizing the result level fusion.

\section{Experiments}

We evaluate our multimodal mapping framework using the RS-seq dataset across seven urban intersections. The comparative analysis assesses mapping accuracy through mIoU metrics, benchmarking against single-modality approaches and evaluating performance across diverse intersection scenarios.

\subsection{Implementation Details}
% In this study, we utilized a manual approach to align the timestamps of the captured images and point cloud data, ensuring that each frame of point cloud data corresponds to an image captured at the same time, thereby guaranteeing data quality. As shown in Fig.~\ref{fig:2}, we apply a point cloud segmentation algorithm to extract the pavement point cloud and subdivide it into a grid. The average intensity value of the point cloud within each grid cell is then used to generate a grayscale image. Next, the image segmentation algorithm identifies different pavement elements, and the pavement point cloud data are projected onto the image based on the internal and external parameters of the imaging equipment to obtain the point cloud data corresponding to the elemental regions. Subsequently, we fuse the point cloud data derived from both the point cloud and image sides, performing denoising and clustering to yield accurate point cloud data for the pavement elements. Finally, we conduct vectorization operations based on the refined point cloud data. After denoising and clustering the linear point cloud data, such as stop lines and lane dividers, we compute the fitted straight line for each point cloud cluster using the least squares fitting algorithm. For surface elements, such as pedestrian crossings, we employ the $\alpha$-shape method to identify and extract the edge points.

In this study, we time-align the captured images and point cloud data with the error controlled within 0.02 seconds to ensure that each frame of point cloud data corresponds to the simultaneously captured images, thus providing the data quality. As shown in Fig.~\ref{fig:2}, we use the point cloud segmentation algorithm to extract the ground point cloud data and perform the gridding process, with the grid edge length set to 0.01 meters. The average intensity value of the point cloud within the grid is utilized to generate an intensity image. Near-neighbour clustering is performed on the segmented point cloud data, and the neighbourhood threshold is set to 0.5 meters. The segmentation results are compared with the ground point cloud data. In the fusion stage of the image segmentation results with ground point cloud data, we used a manually calibrated camera and the calibration parameters of LiDAR to ensure the accuracy of the data.

% We implemented our baseline methods and the benchmark models (U-Net2, PidNet2, ViT-Adapter2) using the PyTorch framework3. Key training hyperparameters were kept consistent across compared models for fairness, unless otherwise specified. Models were trained for 100 epochs using a batch size of 2 per GPU. We employed the AdamW optimizer4 with an initial learning rate set to 1e-4, using standard beta values ($\beta$1=0.9, $\beta$2=0.999) and a weight decay of 0.01. A step decay schedule reducing the learning rate by a factor of 10 at epochs 60 and 90 was used to adjust the learning rate during training. Input images were resized to 512x512 pixels. Point clouds were processed as described in Section 3. All experiments were conducted on NVIDIA RTX 3090 GPUs.

Using the PyTorch framework, we implemented our baseline methods and benchmark models (U-Net, PidNet, and ViT-Adapter). To ensure fairness in comparison, we kept key training hyperparameters consistent across all models unless otherwise specified. Input images were resized to 512x512 pixels, and each model was trained for 100 epochs with a batch size of 2 per GPU. Adam optimizer was used with an initial learning rate of 1e-4, standard beta values $\beta$1=0.9, $\beta$2=0.999), and a weight decay of 0.01. A step decay schedule was applied to reduce the learning rate by a factor of 10 at epochs 60 and 90. All experiments were performed on NVIDIA RTX 3090 GPUs.

Due to the limited amount of data collected from the intersections by the roadside units, training the deep learning network model solely on these seven intersections proves inadequate. Consequently, we used point cloud data and images from NavInfo Company to train the deep learning model, comprising 60,000 images and 20,000 frames of point cloud data. The training data cover typical urban and suburban scenarios, with varying weather conditions. However, the lack of publicly available diverse datasets remains a challenge. The RS-seq dataset was used as a test to validate the experimental method's feasibility.

\subsection{Evaluation Metrics}
\subsubsection{Intersection over Union} The Intersection over Union (IoU) is a widely recognized metric for assessing the accuracy of object detection and segmentation tasks. In the realm of semantic HD map prediction, the IoU between the predicted semantic HD map (${E_1}$) and the ground-truth semantic HD map (${E_2}$) serve as a vital indicator of the expected map’s quality. It quantifies the extent of overlap between the two maps, thereby assessing how well the predicted map corresponds to the ground-truth map. The IoU is formally expressed as follows:

\begin{equation}\label{eq:4}
IoU({E_1},{E_2}) = \frac{{\left| {{E_1} \cap {E_2}} \right|}}{{\left| {{E_1} \cup {E_2}} \right|}}
\end{equation}

\subsubsection{One-way Chamfer Distance} The one-way Chamfer distance (CD) quantifies the similarity between predicted and ground-truth curves. It computes the average distance from each point on the predicted curve to its nearest corresponding point on the ground-truth curve. This metric is mathematically expressed as follows:

\begin{equation}\label{eq:5}
CD{\rm{ = }}\frac{{\rm{1}}}{{{C_1}}}\sum\limits_{x \in {C_1}} {\mathop {\min }\limits_{y \in {C_2}} {{\left\| {x - y} \right\|}_2}}
\end{equation}
where ${C_1}$ and ${C_2}$ denote point sets on the predicted and ground-truth curves. CD assesses spatial distances between curves. However, using CD alone for semantic HD map evaluation has a flaw: smaller IoU tends to result in smaller CD. Thus, we combine CD and IoU to select true positives for better semantic HD map generation assessment.

\subsubsection{Average Precision} The average precision (AP) measures the instance detection capability and is defined as:

\begin{equation}\label{eq:6}
AP = \frac{1}{{10}}\sum\limits_{r \in \left\{ {0.1,0.2,...,1.0} \right\}} {A{P_r}}
\end{equation}
where ${AP_r}$ denotes precision at recall (r). Following the methodology introduced in~\cite{li2022hdmapnet}, the CD is used to identify actual positive instances. Additionally, we will introduce an IoU threshold. An instance is classified as a true positive only if the CD is below a defined threshold and the IoU exceeds the specified threshold. We set the IoU threshold to 0.1 and the CD threshold to 1.0 m.

\subsection{Evaluation Results}

% We evaluate our method on the roadside units dataset - RS-seq dataset, where we focus on the task of considering semantic HD map segmentation and instance detection with three static map elements including lane dividers, pedestrian crossing and stop lines. A comparison of the mIoU scores for semantic map segmentation is given in Table~\ref{tab:1}, where our multimodal achieves the best results for all elements and shows significant improvement at all intervals, which demonstrates the superiority of our approach. From the results, we can observe that Camera-LiDAR fusion methods are usually superior to Camera-only or LiDAR-only methods.

% From Fig.~\ref{fig:2}, it can be seen that the results obtained from the image plus high-intensity point cloud have a better mapping effect for targets at long distances, but their mapping accuracy is poor, and the range of pedestrian crossings is obviously wider. The results obtained from the single point cloud ground intensity map have poorer target mapping effect at the long distance, but it can get more complete target mapping effect at the close distance.

% In Table~\ref{tab:2} shows the results of the comparison test we did using three image segmentation network models U-net~\cite{ronneberger2015u}, Pidnet~\cite{xu2023pidnet} and ViT-Adapter~\cite{chenvision}, we found that for the roadside data, compared to the convolutional neural network based model the transform based Vit-adapter model achieves the best performance on several elements.

We evaluate our method using the RS-seq dataset, focusing on semantic HD map segmentation and instance detection for three static map elements: lane dividers, pedestrian crossings, and stop lines. A comparison of the mIoU scores for semantic map segmentation is presented in Table~\ref{tab:2}. Our multimodal approach achieves the best results across all elements, highlighting the superiority of our method. The results indicate that Camera-LiDAR fusion methods typically outperform Camera-only or LiDAR-only approaches. As presented in Table~\ref{tab:1}, we found through comparative tests on three intersections selected from the RS-seq dataset that the multimodal mapping method delivers superior accuracy in detecting long-range targets.
In contrast, mapping based on a single image exhibits significantly lower accuracy, with pedestrian crosswalks appearing notably wider. Similarly, results derived from a single-point cloud ground intensity map show poor performance in long-range target mapping. However, this approach provides more comprehensive mapping for targets at close distances. In comparison, the multimodal approach achieves more precise detection of pedestrian crosswalks and stop lines at long distances while maintaining high accuracy in their extraction at closer distances. Table~\ref{tab:2} displays the results of our comparative analysis using three image segmentation network models: U-Net~\cite{ronneberger2015u}, PidNet~\cite{xu2023pidnet}, and ViT-Adapter~\cite{chenvision}. The study demonstrates that the transformer-based ViT-Adapter model exceeds the performance of convolutional neural network (CNN) models, especially in the context of roadside data.

\begin{table}[htp!]
\caption{Comparison of mIoU Performance for Different Image Segmentation Models Across Modalities.}
\begin{center}
\begin{threeparttable}
\begin{tabular}{l|l|rrr}
\toprule
{\textbf{Modality}} & {\textbf{Model}} & {\textbf{Ped.}} & {\textbf{St.}} & {\textbf{Div.}}                        \\
\midrule
%Image-seg & \multirow{3}{U-net}{*lane_divider} & \multirow{3}{DeepLab-V3}{*} & \multirow{3}{PIDNet}{*} \\
~\multirow{3}{*}{\textbf{Image}} & U-net~\cite{ronneberger2015u}  & 0.356 & 0.157 & 0.216 \\ 
                           & Pidnet~\cite{xu2023pidnet} & 0.348 & 0.324 & 0.532 \\
                           & ViT-A~\cite{chenvision} & 0.364 & 0.383 & 0.545 \\
\midrule
~\multirow{3}{*}{\textbf{PointCloud}} & U-net~\cite{ronneberger2015u} & 0.206 & 0.103 & 0.265 \\ 
                           & Pidnet~\cite{xu2023pidnet} & 0.186 & 0.213 & 0.407 \\
                           & ViT-A~\cite{chenvision} & 0.229 & 0.224 & 0.433 \\
\midrule
~\multirow{3}{*}{\textbf{Multimodal}}& U-net~\cite{ronneberger2015u} & 0.386 & 0.325 & 0.102 \\ 
                           & Pidnet~\cite{xu2023pidnet} & 0.368 & 0.389 & 0.356 \\
                           & ViT-A~\cite{chenvision} & 0.402 & 0.422 & 0.592 \\ 
\bottomrule
\end{tabular}
\begin{tablenotes}
\scriptsize
\item {Ped.:Pedestrian crossing, St.:Stop line, Div.: Lane divider.}
\end{tablenotes}
\end{threeparttable}
\end{center}
\label{tab:2}
\end{table}

% \begin{figure}
%   \includegraphics[width=\columnwidth]{figures/Fusion_result.jpg}
%   \vspace{-5mm}
%   \caption{Image and point cloud data fusion into a mapping process is presented, and the results of multimodal data fusion are significantly better than the unimodal results.}
%   \label{fig:1}
%     \vspace{-5mm}
% \end{figure}

\begin{figure*}
  \includegraphics[width=\textwidth]{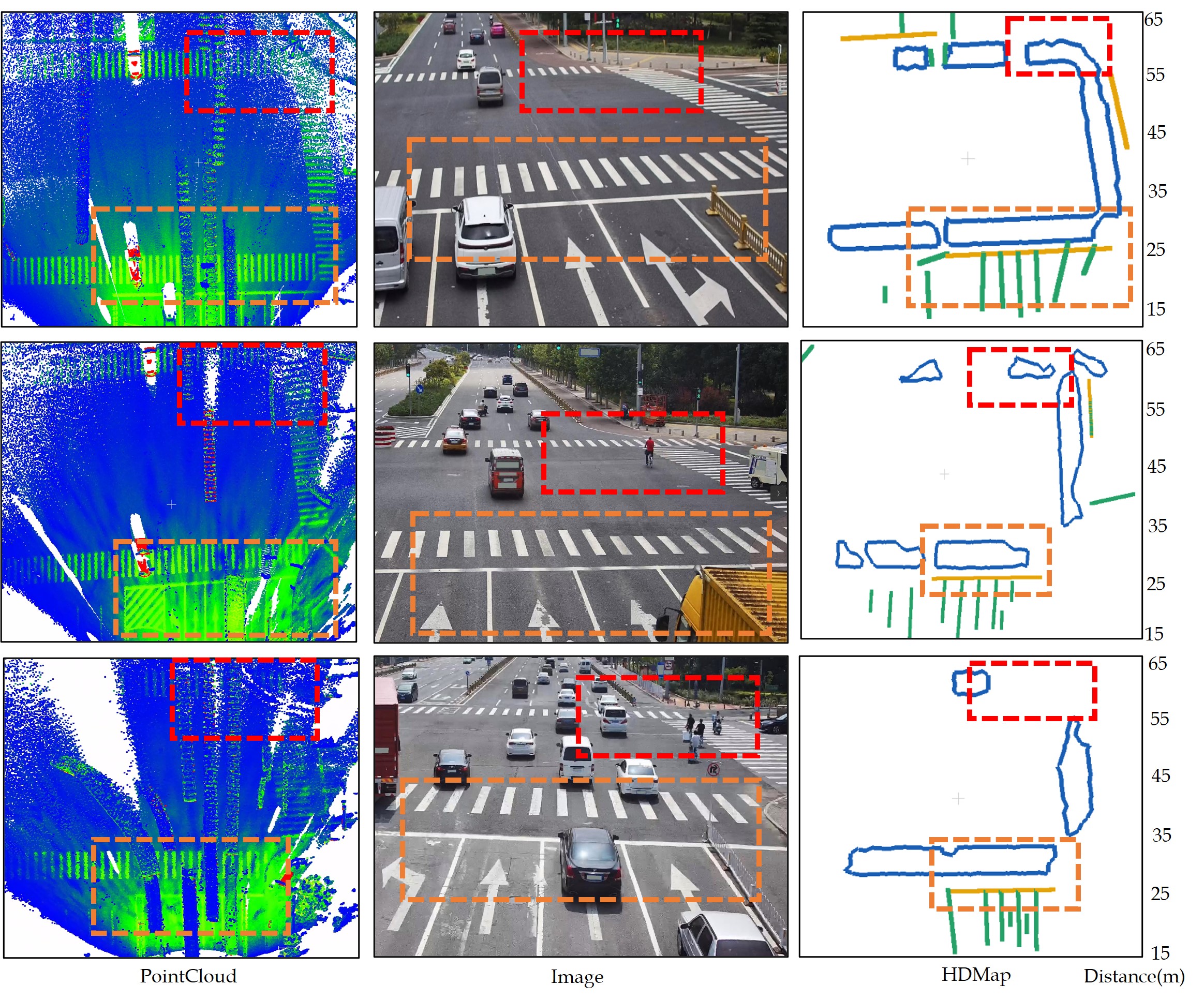}
  
  \caption{The impact of varying distances to sensors on map quality.}
  \label{fig:3}
  
\end{figure*}

\section{Discussions}
 \subsection{Module Selection Analysis}
% Table~\ref{tab:1} illustrates that the multimodal results significantly outperform the unimodal results. It is evident that the actual position and size of the elements deviate from the results obtained from images; for instance, the position of the stop line is notably misaligned, and the width of the pedestrian crossings is substantially broader than desired. This discrepancy indicates that the accuracy of image-based results does not meet the necessary standards due to challenges in the image formation process. In addition, unimodal results derived from point cloud data reveal that elements are missing in areas far from the sensor. The quality of the point cloud map decreases with distance from the sensor, as the intensity values of point cloud data points are affected by environmental interference. This results in insufficient texture clarity for each element, ultimately leading to suboptimal segmentation outcomes. Experimental results show that the mIoU of multimodal recognition improves by an average of approximately 11\% compared to unimodal recognition.

Table~\ref{tab:1} shows the effectiveness of multimodal and unimodal extracting of stop lines, lane markings, and crosswalks at three intersections. The mIoU values of the three elements in the three intersections are given. The results show that the multimodal results are significantly better than the unimodal results. The actual positions and sizes of the elements differ from the results obtained from the images. For example, the parking lines are misplaced, and the width of the pedestrian crossings is much broader than expected. Such discrepancies indicate that the accuracy of the image-based results does not meet the necessary standards due to challenges in the image formation process.
Additionally, the unimodal results obtained from the point cloud data indicate missing elements in areas away from the sensor. The quality of the point cloud image degrades with distance from the sensor because environmental disturbances influence the intensity values of the point cloud data points. This leads to a lack of texture clarity for each element, ultimately leading to sub-optimal segmentation results. Experimental results show an average improvement of about 11\% in mIoU for multimodal compared to unimodal recognition.

In our experimental evaluation, we initially benchmarked several established image segmentation architectures to assess their suitability for processing roadside sensor data within our multimodal framework. As presented in Table II, we compared U-Net, PidNet, and ViT-Adapter using image-only, point cloud-only, and multimodal fusion approaches. The results indicated that the Transformer-based ViT-Adapter generally achieved favourable performance on several map elements compared to the CNN-based models for this specific task and dataset. Furthermore, the multimodal fusion approach consistently exceeded single-modality inputs, highlighting the benefit of combining camera and LiDAR data from RSUs.

% We acknowledge that the field of HD mapping has rapidly evolved, with methods like HDMapNet, BEVFusion, and SuperFusion representing the current state-of-the-art for online, vehicle-based map generation. A direct quantitative comparison between our RSU-based approach and these vehicle-centric SOTA methods on a unified benchmark is non-trivial due to fundamental differences in the sensor configurations, viewpoints, data characteristics , and the specific focus (intersection mapping vs. general road mapping). Adapting and retraining these complex models, originally designed for vehicle platforms, on our RS-seq dataset6 would require significant effort and lies beyond the scope of this initial study but constitutes an important avenue for future work.

\begin{table}[ht]
\caption{Comparison Of mIoU Performance Of Different End-to-end Vectorized semantic HD Map Models On Roadside Datasets }
\begin{center}
\begin{tabular}{l|rrrrr}
\toprule
{\textbf{Model}} & {\textbf{Ped.}} & {\textbf{St.}} & {\textbf{Div.}}  \\ 
\midrule
\textbf{BEVFusiont~\cite{liang2022bevfusion}} & 0.163  & 0.225 & 0.047 \\  
\midrule
\textbf{HDMapNet~\cite{li2022hdmapnet}} & 0.218  & 0.242 & 0.164  \\  
\midrule
\textbf{SuperFusion~\cite{10611320}} & 0.232  & 0.014 & 0.135 \\  
\bottomrule
\end{tabular}
\end{center}

\label{tab:3}
\end{table}

The results presented in Table~\ref{tab:3} indicate that when three end-to-end mapping methods are directly applied to the roadside dataset, their performance is suboptimal, as evidenced by generally low overall mIoU scores. Analysis reveals that significant discrepancies exist in the data acquisition perspectives between vehicle-based data, potentially used for pre-training, and the roadside data employed in this study. The generalization ability is constrained due to the limited ability of the model to adjust to the feature distribution of the new dataset accurately. To improve the model’s performance in roadside scenarios, researchers urgently need to expand the size of the roadside dataset further and carry out end-to-end training of the model to enhance its ability to capture the features of the roadside scenarios and improve its generalization performance.
% Therefore, the comparisons presented in Table II4 primarily serve to establish strong baselines within the context of RSU-based multimodal mapping on our proposed RS-seq dataset6. They validate the effectiveness of multimodal fusion in this setting and provide an initial assessment of different network backbones5. Our main contribution lies in proposing and evaluating a framework tailored for leveraging the unique advantages of RSU infrastructure for detailed intersection mapping7, complementing existing vehicle-based mapping paradigms. Future work will involve comparative studies against adapted state-of-the-art online mapping methods to further contextualize the performance of RSU-based systems.

\subsection{Frame Rate Impact on Map Quality}
Multi-frame point cloud data greatly enhances detail capture in creating semantic HD maps by increasing data density and coverage, producing a more detailed and refined final mapping output. At intersections, multi-frame point clouds and image data facilitate the continuous capture of map elements over time, effectively addressing occlusions caused by pedestrians and vehicles. Although multi-frame point cloud and image data are crucial for enhancing mapping quality and ensuring that 3D reconstructions closely resemble real-world scenes, data processing efficiency decreases as the data volume increases, posing greater computational demands on roadside units. 
Table~\ref{tab:4} shows the mIoU scores and computing times for different frame counts in multimodal experiments. As the number of frames increases, the mIoU score improves, but the computing time also rises significantly. At 50 frames, the mIoU score is sufficiently high, and the computing time remains manageable, achieving the best balance between quality and efficiency.
% We present the results regarding map quality and processing efficiency as the number of frames increases in Table~\ref{tab:3}, illustrating that an optimal balance between map quality and efficiency is achieved at 50 frames.

% \begin{figure}
%   \includegraphics[width=\columnwidth]{figures/fig_4.jpg}
%   \vspace{-5mm}
%   \caption{Image and point cloud data fusion into a mapping process is presented, and the results of multimodal data fusion are significantly better than the unimodal results.}
%   \label{fig:1}
%     \vspace{-5mm}
% \end{figure}

% \begin{table}[ht]
% \caption{IoU scores of HD map on different frames data}
% \begin{center}
% \begin{tabular}{l|rrrrr}
% \toprule
% \textbf{\diagbox{Method}{Frame}} & {\textbf{1}} & {\textbf{10}} & {\textbf{20}} & {\textbf{50}} & {\textbf{100}}  \\
% \midrule
% ~Image & 0.251 & 0.304 & 0.392 & 0.431 & 0.455 \\        
% \midrule
% ~PointCloud & 0.210  & 0.231 & 0.251 & 0.292 & 0.317 \\      
% \midrule
% ~Multimodal & 0.268  & 0.324 & 0.423 & 0.475 & 0.498 \\  
% \bottomrule
% \end{tabular}
% \end{center}

% \label{tab:3}
% \end{table}

\begin{table}[ht]
\caption{Comparison of mIoU Scores and Computing Times Across Different Frame Counts in Multimodal Experiments}
\begin{center}
\begin{tabular}{l|rrrrr}
\toprule

\textbf{Frame} & {\textbf{1}} & {\textbf{10}} & {\textbf{20}} & {\textbf{50}} & {\textbf{100}}  \\   
\midrule
\textbf{mIoU} & 0.268  & 0.324 & 0.423 & 0.475 & 0.498 \\  
\midrule
\textbf{Computing time(s)} & 0.321  & 3.124 & 5.432 & 15.378 & 33.459 \\  

\bottomrule
\end{tabular}
\end{center}

\label{tab:4}
\end{table}

\subsection{Impact of Object-Sensor Distance on Map Quality}
The placement of the camera and LiDAR on the same side of the intersection allows for an intuitive comparison of the effects of distance on point cloud and image mapping. The distance between the sensor and the object primarily affects resolution and detail capture. Shorter distances enable sensors to capture finer surface details, resulting in higher point cloud density and increased image resolution, producing more accurate and detailed mapping outcomes. As shown in Fig.~\ref{fig:3}, as the distance between the object and the sensor increases, the density of the point cloud data gradually decreases. Furthermore, objects located farther from the camera’s centre exhibit imaging aberrations, which can degrade map quality. Table~\ref{tab:5} illustrates that the number of practical points per square meter progressively decreases as distance increases, demonstrating the impact of distance on mapping accuracy.

% \begin{table}[h]
% \caption{Influence of Distance on Mapping Effect.}
% \begin{center}
% \begin{tabular}{c|c|c|c|c}
% \toprule
% {\textbf{Distance}} & {\textbf{PointCloud}} & {\textbf{Image}} & {\textbf{HDMap}} & {\textbf{mIoU}}                                       \\ 
% \midrule
% \makecell[c]{15-25} & \includegraphics[width=0.5in]{figures/Diss-Image/p1.png}  & \includegraphics[width=0.5in]{figures/Diss-Image/i1.png} & \includegraphics[width=0.5in]{figures/Diss-Image/h1.png} & 0.89 \\
% \midrule
% \makecell[c]{25-35} & \includegraphics[width=0.5in]{figures/Diss-Image/p2.png}  & \includegraphics[width=0.5in]{figures/Diss-Image/i2.png} & \includegraphics[width=0.5in]{figures/Diss-Image/h2.png} & 0.80 \\
% \midrule
% \makecell[c]{35-45} & \includegraphics[width=0.5in]{figures/Diss-Image/p3.png}  & \includegraphics[width=0.5in]{figures/Diss-Image/i3.png} & \includegraphics[width=0.5in]{figures/Diss-Image/h3.png} & 0.91 \\
% \midrule
% \makecell[c]{45-55} & \includegraphics[width=0.5in]{figures/Diss-Image/p4.png}  & \includegraphics[width=0.5in]{figures/Diss-Image/i4.png} & \includegraphics[width=0.5in]{figures/Diss-Image/h4.png} & 0.54 \\
% \bottomrule
% \end{tabular}
% %}
% \end{center}
% \label{tab:4}
% \end{table}

\begin{table}[ht]
\caption{The influence of Sensor Distance on points density and map quality.}
\begin{center}
\begin{tabular}{l|rrrrr}
\toprule
\textbf{Distance (m)} & \textbf{15-25}  & \textbf{25-35} & \textbf{35-45} & \textbf{45-55} & \textbf{55-65 }\\        \midrule
\textbf{PointDensity (N/${m^2}$)} & 54  & 42 & 25 & 20 & 11 \\ 
\midrule
\textbf{mIoU} & 0.891 & 0.802 & 0.742 & 0.546 & 0.213 \\
\bottomrule
\end{tabular}
\end{center}
\label{tab:5}
\end{table}

\section{Conclusion}
% In this study, we present the RS-seq dataset, which is based on the V2X-seq dataset to refine its content in the process of creating HD maps. This dataset is distinguished by its inclusion of detailed annotations corresponding to various road features. Building upon the RS-seq dataset, we propose a novel approach for leveraging data from intelligent roadside units at intersections to enhance map quality. Specifically, we present a baseline method for generating HD maps utilizing multimodal data from these roadside units. The experimental results demonstrate that the integration of intelligent roadside unit data significantly improves the completeness and accuracy of HD maps. These findings underscore the potential of intelligent roadside units to advance the field of map production and contribute to the development of more precise and reliable autonomous driving systems.

%

This paper introduces a comprehensive framework for HD semantic map generation leveraging intelligent roadside units, demonstrating notable improvements over traditional vehicle-based methods. Our multimodal fusion methodology achieved a 4\% increase in mIoU compared to the image-only baseline and an 18\% increase compared to the point cloud-only baseline, particularly improving accuracy in challenging scenarios, such as occluded road elements and the creation of semantic HD maps for urban intersections. Additionally, we developed and released the RS-seq dataset, which comprises precisely labelled camera imagery and LiDAR point clouds collected from seven complex intersections. This dataset addresses a critical gap in research resources for infrastructure-assisted automatic mapping systems. The dataset includes detailed annotations of lane dividers, pedestrian crossings, and stop lines, enabling systematic evaluation of mapping methodologies under realistic conditions. Finally, The study acknowledges limitations, such as computational overhead, performance degradation with increasing distance, and the potential risk of overfitting due to limited public dataset diversity. Future work will focus on developing more robust data augmentation techniques and exploring advanced pretraining methods to enhance model generalization.

\balance
\bibliographystyle{IEEEtran}
\bibliography{IEEEexample}

\begin{thebibliography}{10}
\providecommand{\url}[1]{#1}
\csname url@rmstyle\endcsname
\providecommand{\newblock}{\relax}
\providecommand{\bibinfo}[2]{#2}
\providecommand\BIBentrySTDinterwordspacing{\spaceskip=0pt\relax}
\providecommand\BIBentryALTinterwordstretchfactor{4}
\providecommand\BIBentryALTinterwordspacing{\spaceskip=\fontdimen2\font plus
\BIBentryALTinterwordstretchfactor\fontdimen3\font minus \fontdimen4\font\relax}
\providecommand\BIBforeignlanguage[2]{{%
\expandafter\ifx\csname l@#1\endcsname\relax
\typeout{** WARNING: IEEEtran.bst: No hyphenation pattern has been}%
\typeout{** loaded for the language `#1'. Using the pattern for}%
\typeout{** the default language instead.}%
\else
\language=\csname l@#1\endcsname
\fi
#2}}

\bibitem{li2022hdmapnet}
Q.~Li, Y.~Wang, Y.~Wang, and H.~Zhao, ``Hdmapnet: An online hd map construction and evaluation framework,'' in \emph{2022 International Conference on Robotics and Automation (ICRA)}.\hskip 1em plus 0.5em minus 0.4em\relax IEEE, 2022, pp. 4628--4634.

\bibitem{liao2023maptrv2}
B.~Liao, S.~Chen, Y.~Zhang, B.~Jiang, Q.~Zhang, W.~Liu, C.~Huang, and X.~Wang, ``Maptrv2: An end-to-end framework for online vectorized hd map construction,'' \emph{arXiv preprint arXiv:2308.05736}, 2023.

\bibitem{liu2024mgmap}
X.~Liu, S.~Wang, W.~Li, R.~Yang, J.~Chen, and J.~Zhu, ``Mgmap: Mask-guided learning for online vectorized hd map construction,'' in \emph{Proceedings of the IEEE/CVF Conference on Computer Vision and Pattern Recognition}, 2024, pp. 14\,812--14\,821.

\bibitem{yu2023v2x}
H.~Yu, W.~Yang, H.~Ruan, Z.~Yang, Y.~Tang, X.~Gao, X.~Hao, Y.~Shi, Y.~Pan, N.~Sun, \emph{et~al.}, ``V2x-seq: A large-scale sequential dataset for vehicle-infrastructure cooperative perception and forecasting,'' in \emph{Proceedings of the IEEE/CVF Conference on Computer Vision and Pattern Recognition}, 2023, pp. 5486--5495.

\bibitem{lee2020design}
J.~Lee, K.~Lee, A.~Yoo, and C.~Moon, ``Design and implementation of edge-fog-cloud system through hd map generation from lidar data of autonomous vehicles,'' \emph{Electronics}, vol.~9, no.~12, p. 2084, 2020.

\bibitem{Lang_2019_CVPR}
A.~H. Lang, S.~Vora, H.~Caesar, L.~Zhou, J.~Yang, and O.~Beijbom, ``Pointpillars: Fast encoders for object detection from point clouds,'' in \emph{Proceedings of the IEEE/CVF Conference on Computer Vision and Pattern Recognition (CVPR)}, June 2019.

\bibitem{liang2022bevfusion}
T.~Liang, H.~Xie, K.~Yu, Z.~Xia, Z.~Lin, Y.~Wang, T.~Tang, B.~Wang, and Z.~Tang, ``Bevfusion: A simple and robust lidar-camera fusion framework,'' \emph{Advances in Neural Information Processing Systems}, vol.~35, pp. 10\,421--10\,434, 2022.

\bibitem{philion2020lift}
J.~Philion and S.~Fidler, ``Lift, splat, shoot: Encoding images from arbitrary camera rigs by implicitly unprojecting to 3d,'' in \emph{Computer Vision--ECCV 2020: 16th European Conference, Glasgow, UK, August 23--28, 2020, Proceedings, Part XIV 16}.\hskip 1em plus 0.5em minus 0.4em\relax Springer, 2020, pp. 194--210.

\bibitem{yan2018second}
Y.~Yan, Y.~Mao, and B.~Li, ``Second: Sparsely embedded convolutional detection,'' \emph{Sensors}, vol.~18, no.~10, p. 3337, 2018.

\bibitem{10611320}
H.~Dong, W.~Gu, X.~Zhang, J.~Xu, R.~Ai, H.~Lu, J.~Kannala, and X.~Chen, ``Superfusion: Multilevel lidar-camera fusion for long-range hd map generation,'' in \emph{2024 IEEE International Conference on Robotics and Automation (ICRA)}, 2024, pp. 9056--9062.

\bibitem{xu2022v2x}
R.~Xu, H.~Xiang, Z.~Tu, X.~Xia, M.-H. Yang, and J.~Ma, ``V2x-vit: Vehicle-to-everything cooperative perception with vision transformer,'' in \emph{European conference on computer vision}.\hskip 1em plus 0.5em minus 0.4em\relax Springer, 2022, pp. 107--124.

\bibitem{Ye_2022_CVPR}
X.~Ye, M.~Shu, H.~Li, Y.~Shi, Y.~Li, G.~Wang, X.~Tan, and E.~Ding, ``Rope3d: The roadside perception dataset for autonomous driving and monocular 3d object detection task,'' in \emph{Proceedings of the IEEE/CVF Conference on Computer Vision and Pattern Recognition (CVPR)}, June 2022, pp. 21\,341--21\,350.

\bibitem{yu2022dair}
H.~Yu, Y.~Luo, M.~Shu, Y.~Huo, Z.~Yang, Y.~Shi, Z.~Guo, H.~Li, X.~Hu, J.~Yuan, \emph{et~al.}, ``Dair-v2x: A large-scale dataset for vehicle-infrastructure cooperative 3d object detection,'' in \emph{Proceedings of the IEEE/CVF Conference on Computer Vision and Pattern Recognition}, 2022, pp. 21\,361--21\,370.

\bibitem{zimmer2024tumtrafv2x}
W.~Zimmer, G.~A. Wardana, S.~Sritharan, X.~Zhou, R.~Song, and A.~C. Knoll, ``Tumtraf v2x cooperative perception dataset,'' in \emph{2024 IEEE/CVF Conference on Computer Vision and Pattern Recognition, CVPR}.\hskip 1em plus 0.5em minus 0.4em\relax IEEE, 2024.

\bibitem{he2023vi}
Y.~He, C.~Bian, J.~Xia, S.~Shi, Z.~Yan, Q.~Song, and G.~Xing, ``Vi-map: Infrastructure-assisted real-time hd mapping for autonomous driving,'' in \emph{Proceedings of the 29th Annual International Conference on Mobile Computing and Networking}, 2023, pp. 1--15.

\bibitem{dosovitskiy2017carla}
A.~Dosovitskiy, G.~Ros, F.~Codevilla, A.~Lopez, and V.~Koltun, ``Carla: An open urban driving simulator,'' in \emph{Conference on robot learning}.\hskip 1em plus 0.5em minus 0.4em\relax PMLR, 2017, pp. 1--16.

\bibitem{2020Denoising}
J.~Guo, W.~Feng, T.~Hao, P.~Wang, and H.~Mao, ``Denoising of a multi-station point cloud and 3d modeling accuracy for substation equipment based on statistical outlier removal,'' in \emph{2020 IEEE 4th Conference on Energy Internet and Energy System Integration (EI2)}, 2020.

\bibitem{1994Three}
H.~Edelsbrunner and E.~P. Mücke, ``Three-dimensional alpha shapes,'' \emph{ACM Transactions on Graphics}, vol.~13, no.~1, 1994.

\bibitem{york1966least}
D.~York, ``Least-squares fitting of a straight line,'' \emph{Canadian Journal of Physics}, vol.~44, no.~5, pp. 1079--1086, 1966.

\bibitem{ronneberger2015u}
O.~Ronneberger, P.~Fischer, and T.~Brox, ``U-net: Convolutional networks for biomedical image segmentation,'' in \emph{Medical image computing and computer-assisted intervention--MICCAI 2015: 18th international conference, Munich, Germany, October 5-9, 2015, proceedings, part III 18}.\hskip 1em plus 0.5em minus 0.4em\relax Springer, 2015, pp. 234--241.

\bibitem{xu2023pidnet}
J.~Xu, Z.~Xiong, and S.~P. Bhattacharyya, ``Pidnet: A real-time semantic segmentation network inspired by pid controllers,'' in \emph{Proceedings of the IEEE/CVF conference on computer vision and pattern recognition}, 2023, pp. 19\,529--19\,539.

\bibitem{chenvision}
Z.~Chen, Y.~Duan, W.~Wang, J.~He, T.~Lu, J.~Dai, and Y.~Qiao, ``Vision transformer adapter for dense predictions,'' in \emph{The Eleventh International Conference on Learning Representations}.

\end{thebibliography}
\end{document}